\documentclass[letterpaper, 10pt, conference, final]{ieeeconf}

\IEEEoverridecommandlockouts 

\usepackage[font=small]{caption}
\usepackage{subcaption}
\usepackage{graphicx}
\usepackage{tikz}
\usepackage{xcolor}

\usepackage{amsthm}
\usepackage{amsfonts}
\usepackage{amsmath}
\usepackage{amssymb}

\usepackage{algorithm}
\usepackage{algorithmic}
\usepackage{float}

\makeatletter
\let\NAT@parse\undefined
\makeatother
\usepackage{url}
\usepackage[colorlinks,citecolor=blue,linkcolor=red]{hyperref}
\usepackage{cite}
\usepackage{cleveref}
\crefrangeformat{equation}{(#3#1#4)--(#5#2#6)}

\newtheorem{theorem}{Theorem}
\newtheorem{definition}{Definition}

\newcommand{\R}{\mathbb{R}}

\newcommand{\T}{^\intercal}

\title{\bf Distributed Potential iLQR: Scalable Game-Theoretic Trajectory Planning for Multi-Agent Interactions}

\author{Zach Williams$^1$, Jushan Chen$^2$, Negar Mehr$^2$ 
\thanks{This work is supported by the National Science Foundation, under grants ECCS-2145134 CAREER Award, CNS-2218759, and CCF-2211542.}
    \thanks{$^1$Zach Williams is with the Department of Electrical and Computer
        Engineering, University of Illinois Urbana-Champaign, 306 N Wright St, Urbana,
        IL 61801, USA, {\tt\small zjw4@illinois.edu} as well as Raytheon BBN Technologies 
        in the Network \& Cyber Technologies Group, 10 Moulton Street, Cambridge, MA 02138, USA}
    \thanks{$^2$Jushan Chen and Negar
        Mehr are with the Department of Aerospace Engineering, University of Illinois
        Urbana-Champaign, 104 S Wright St, Urbana, IL 61801, USA, {\tt\small \{jushanc2,
        negar\}@illinois.edu}}
}

\begin{document}

\maketitle
\thispagestyle{empty}
\pagestyle{empty}

%%%%%%%%%%%%%%%%%%%%%%%%%%%%%%%%%%%%%%%%%%%%%%%%%%%%%%%%%%%%%%%%%%%%%%%%%%%%%%%%%%%%%%%
\begin{abstract}

In this work, we develop a scalable, local trajectory optimization algorithm that
enables robots to interact with other robots. It has been shown that agents'
interactions can be successfully captured in game-theoretic formulations, where the
interaction outcome can be best modeled via the equilibria of the underlying dynamic
game. However, it is typically challenging to compute equilibria of dynamic games as it
involves simultaneously solving a set of coupled optimal control problems. Existing
solvers operate in a centralized fashion and do not scale up tractably to multiple
interacting agents. We enable scalable distributed game-theoretic planning by leveraging
the structure inherent in multi-agent interactions, namely, interactions belonging to
the class of dynamic potential games. Since equilibria of dynamic potential games can be
found by minimizing a single potential function, we can apply distributed and
decentralized control techniques to seek equilibria of multi-agent interactions in a
scalable and distributed manner. We compare the performance of our algorithm with a
centralized interactive planner in a number of simulation studies and demonstrate that
our algorithm results in better efficiency and scalability. We further evaluate our
method in hardware experiments involving multiple quadcopters.\footnote{Code Repository-
\href{https://github.com/labicon/dp-ilqr}{\texttt{https://github.com/labicon/dp-ilqr}}}

\end{abstract}

 % Keywords:
 % - dynamic game theory
 % - multi-agent navigation
 % - potential games
 % - Path planning for multiple robots
 % - Multi-robot systems
 % - interactive trajectory planning

%%%%%%%%%%%%%%%%%%%%%%%%%%%%%%%%%%%%%%%%%%%%%%%%%%%%%%%%%%%%%%%%%%%%%%%%%%%%%%%%%%%%%%%
\section{INTRODUCTION} \label{sec:introduction}

\begin{figure*}[ht]
    \centering
    \includegraphics[width=\textwidth]{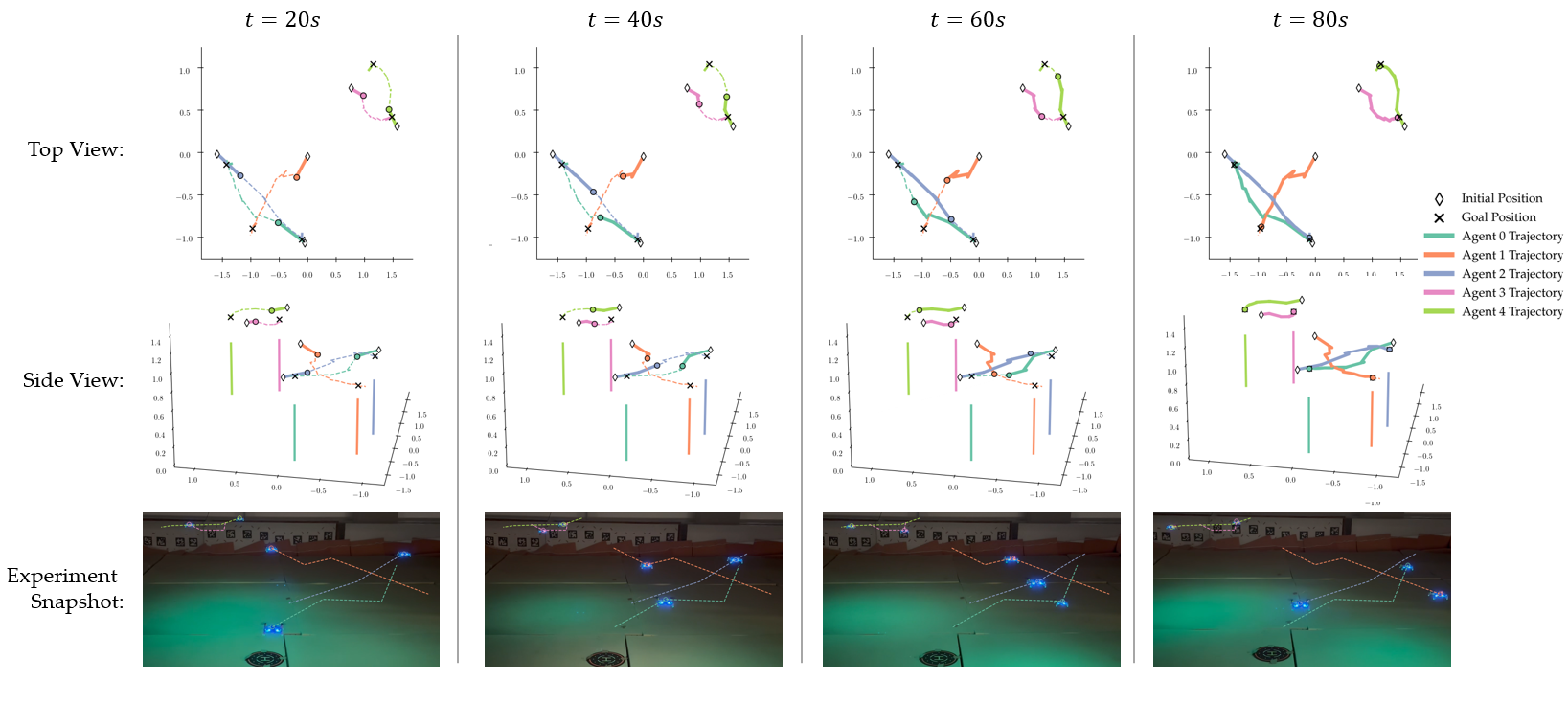}
    \caption{Hardware experiment demonstrated on 
    \href{https://www.bitcraze.io/products/crazyflie-2-1/}{Crazyflie 2.1} using the 
    Crazyswarm library \cite{preiss2017crazyswarm}. In this scenario, five drones 
    navigate around each other in two separate intersections. As time evolves from 
    left to right, the drones are able to safely navigate around each other. The first 
    row depicts the birds-eye view, the middle row depicts the 3D rendered side view, 
    and the lower row shows live pictures of the experiment. The vertical lines 
    correspond to each of the $\times$ markers and denote the $(x, y)$ of the 
    intended goal position colored by agent.}
    \label{fig:5drone-experiment}
    \vspace{-2ex}
\end{figure*}

Automatically generating intuitive trajectories in interactive robotic applications
involving multiple agents is an important and challenging problem~\cite{yang2018grand}.
In situations where scalability matters, we need algorithms that enable robots to safely
and tractably navigate around other agents. For example, crowd robot navigation and
autonomous driving may require a robot to navigate among multiple humans. Alternatively,
drone delivery systems may require quadcopters to navigate by multiple drones in an
aerial space. A critical challenge in trajectory planning in such interactive settings
is that each robot must account for the likely reactions of other agents to its action,
which results in a coupling among the agents, accounting for which quickly becomes
intractable as the number of agents increases. In this work, we address this challenge
and develop a scalable trajectory optimization algorithm that enables robots to interact
efficiently with other robots.

Game-theoretic planning has proven to be a powerful framework for capturing interactions
between independent agents~\cite{sadigh2016planning,wang2019game} where an agent seeks
to maximize its utility. Since agents' utilities depend on the state and actions of all
agents, the interaction outcome can be best represented via equilibria that account for
the mutual influence of agents. While conceptually powerful, the resulting equilibria
are hard to compute as it involves simultaneously solving a set of coupled optimal
control problems. Several recent works have developed approximate trajectory
optimization algorithms for such games~\cite{fridovich2019ilqgames,
laine2021computation, wang2020game}. However, all these works operate in a centralized
fashion, i.e. they require the robot to account for interactions with all agents in the
environment. As a result, they rarely extend beyond three agents, even for simple
dynamics models.

We consider the couplings that result from agents' interactions in a game-theoretical
setup and draw from decentralized and distributed MPC literature to develop a
distributed trajectory planner that can be run in a receding horizon fashion efficiently
and scalably. We leverage the fact that certain multi-agent interactions are part of a
more special class of games, namely dynamic potential games~\cite{kavuncu2021potential}.
Potential games are a class of games for which equilibria can be found efficiently and
tractably by solving a single optimization problem. Our key insight is that since
equilibria of dynamic potential games can be found by minimizing a single potential
function, we can apply techniques from distributed and decentralized model predictive
control to seek Nash equilibria of game-theoretic interactions in a scalable and
efficient manner. 

Our algorithm is distributed because each agent independently computes its control input
using the state information of its neighboring agents. We compare the performance of our
algorithm with a centralized interactive planner in a number of simulation studies and
demonstrate that our algorithm can efficiently account for interactions with a larger
number of agents. We further showcase the success of our method in hardware experiments
involving multiple quadcopters.

The organization of this paper is as follows. In Section~\ref{sec:related-work}, we
review related works. In Section~\ref{sec:problem-formulation}, we introduce the
planning problem to be solved. In Section~\ref{sec:prior-work}, we discuss the previous
work that utilizes dynamic potential games. In Section~\ref{sec:dist-traj-opt}, we
define our distributed trajectory planning algorithm. In
Sections~\ref{sec:simulation-studies} and~\ref{sec:experiments}, we highlight empirical
results in simulations and on hardware. We conclude the paper in
Section~\ref{sec:conclusion}.

%%%%%%%%%%%%%%%%%%%%%%%%%%%%%%%%%%%%%%%%%%%%%%%%%%%%%%%%%%%%%%%%%%%%%%%%%%%%%%%%%%%%%%%
\section{RELATED WORK} \label{sec:related-work}

Reactive methods that utilize multi-modal probabilistic prediction models of agents are
one of the popular approaches to interactive trajectory planning
\cite{schmerling2017multimodal, nishimura2020risk, bai2015intention}. The downside with
these approaches is that agents are not able to sufficiently influence each other,
resulting in conservative interactions. Consequently, several recent methods have
considered game-theoretic planning for interactive domains, which enables agents to
influence one another and achieve joint prediction and planning. Several approaches that
rely on Differential Dynamic Programming have been proposed \cite{sun2015game,
di2018differential, fridovich2019ilqgames} for finding Nash equilibria of general
dynamic games, with \cite{fridovich2019ilqgames} demonstrated in real-time. The
hierarchical method shown in~\cite{fisac2019hierarchical} decomposes the problem into a
strategic global planner and a tactical local planner, but it requires discretizing the
state and action spaces. Dynamic programming approaches were further utilized
in~\cite{wang2020game} and~\cite{mehr2023maximum} to approximately find equilibria of
interactions under uncertainty. In~\cite{cleac2019algames}, equilibria of interactive
dynamic games were sought under nonlinear state and input constraints. Implicit methods
utilize some form of inverse reinforcement learning over trajectory datasets to achieve
collision avoidance without directly imposing constraints on the structure of
interactions \cite{ziebart2009planning, henry2010Crowded, kretzschmar2016socially},
whereas we explicitly take advantage of the structure of certain multi-agent
interactions to simplify the problem.

There has been a myriad of approaches exploring the various forms of game-theoretic
equilibria among agents. Stackelberg equilibria were initially used to model the
outcomes~\cite{yoo2012stackelberg, sadigh2016planning, liniger2019noncooperative}.
However, it proved insufficient for general forms of interactions because it assumes a
leader-follower structure, which does not apply to non-cooperative games with more than
two agents. Due to these challenges, others considered Nash equilibria to capture the
interactions among more than two agents. 

When it comes to computing equilibria, potential games are a class of games for which
equilibria can be found efficiently and reliably. While much of the literature on
potential games is oriented toward the static case, there have been several recent
advancements in dynamic potential games. Following the pioneering works
in~\cite{dechert1978optimal} and~\cite{dechert2006stochastic}, there were initially two
primary methods of solving dynamic games: Euler-Lagrange and Pontryagin's Maximum
methods. More recently, a Hamiltonian potential function was explored
in~\cite{dragone2015hamiltonian} in the case of open-loop games for continuous time
models. While~\cite{zazo2016dynamicPotential} was primarily focused on communications
applications, they demonstrated successful utilization of the simplified problem
structure offered by potential games. Similar to our work,~\cite{dechert1997non} posed
the idea of connecting the open-loop Nash equilibria of a dynamical game with the
solutions to an optimal control problem under certain conditions. Recently,
\cite{kavuncu2021potential} and \cite{bhatt2022efficient} demonstrated that dynamic
potential games can be leveraged for trajectory planning in interactive robotics, but
this used a centralized construction.

%%%%%%%%%%%%%%%%%%%%%%%%%%%%%%%%%%%%%%%%%%%%%%%%%%%%%%%%%%%%%%%%%%%%%%%%%%%%%%%%%%%%%%%
\section{Problem Formulation} \label{sec:problem-formulation}

Consider an interactive trajectory planning problem with $N$ agents. Let $[N]\equiv \{1,
\dots, N\}$ be the set of agents' indices. We refer to the state of the $i$th agent at
time $k$ as $x_k^i \in \R^{n_i}$, where $n_i$ is the dimension of the state vector of
agent $i \in [N]$. Agent $i$'s corresponding control input is denoted as $u_k^i \in
\R^{m_i}$, where $m_i$ is the dimension of the control space of agent $i \in [N]$.
Concatenating the states and inputs of all agents, the full state vector of the system
at time $k$ is given by $x_k = (x^1_k, x^2_k, \dots, x^{N}_k) \in \R^n$, where
$\sum_{i=1}^{N} n_i \equiv n$. The concatenated set of control inputs of all agents at
time $k$ is similarly denoted by $u_k = (u^1_k, u^2_k, \dots, u^{N}_k) \in \R^m$, where
$\sum_{i=1}^{N} m_i \equiv m$. Generally, subscripts are used to denote the time index,
whereas superscripts denote the agent index. The states for agent $i$ across an entire
horizon $T$ is notated as $x^i = \{x_0^i, \dots, x_T^i\} \in \R^{n_i \times T}$.
Similarly, for the controls, let $u^i = \{u_0^i, \dots, u_T^i\} \in \R^{m_i \times T}$
be agent $i$'s control inputs across the horizon. We drop both subscript and superscript
to refer to all agents over an entire horizon for states $x \in \R^{n \times T}$ and
controls $u \in \R^{m \times T}$. Lastly, we use capitalization to differentiate the
predicted states $X$ from the actual states $x$ and the same for the controls $U$ from
$u$, respectively.

We consider separable agents' dynamics defined by $f^i : \R^{n_i} \times \R^{m_i}
\mapsto \R^{n_i}$, i.e. we assume that for every agent $i \in [N]$, we have:
\begin{equation} \label{eqn:decoupled-dynamics}
    x^i_{k+1} = f^i (x^i_k, u^i_k).
\end{equation}
We denote the strategy space of each agent $i \in [N]$ as $\Gamma^i$. Let the strategy
$\gamma^i \in \Gamma^i$ of agent $i$ be given by $
\gamma^i:\mathbb{R}^{n_i}\times\{0,1,\ldots,T-1\}\mapsto \mathbb{R}^{m_i}$ which
determines the actions of agent $i$ at all time instants. We consider open-loop
strategies that are only a function of the system's initial state and the time step.
Therefore, we have $\gamma^i(x_0,k) := u^i_k$. Hence, for simplicity, we use strategies
and actions interchangeably here out.

We assume that each agent $i$ is minimizing some cost $J^i(\cdot)$ over the time horizon
$T$:
\begin{equation}\label{eq:cost}
    J^i(x_0,\gamma) = S^i(x_{T},T) + \sum_{k=0}^{T-1}L^i(x_k,\gamma(x_0,k),k),
\end{equation}
where $S^i(\cdot)$ is the terminal cost of agent $i$ and $L^i(\cdot)$ is the running
cost of agent $i$. Note that the cost perceived by an agent $i$ may depend on the states
and actions of all the other agents. As a result, generally, it is not possible for all
agents to optimize their costs simultaneously, and we need to model the outcome of the
interactions between the agents as equilibria of the underlying dynamic game. We denote
our dynamic games by a compact notation, $G_{x_0}^T := (T, \{\gamma^i\}_{i=1}^N,
\{J^i\}_{i=1}^N, \{f^i\}_{i=1}^N),$ which denotes the dynamic game that arises from
interactions of the agents over a horizon $T$ starting from the initial condition $x_0$.

Let $\gamma^{-i} = (\gamma^1, \dots, \gamma^{i-1}, \gamma^{i+1}, \dots, \gamma^{N})$
denote the strategies of all other agents except $i$. We use similar notation to express
the states and controls of all other agents except $i$ for a given time step $k$ as
$x_k^{-i} \in \R^{n - n_i}$ and $u_k^{-i} \in \R^{m - m_i}$, respectively. Then, the
Nash equilibria of our dynamic game are defined as~\cite{basar1998dynamic}:

\begin{definition}\label{def:Nash} For a given game, $G_{x_0}^T := (T,
\{\gamma^i\}_{i=1}^N, \{J^i\}_{i=1}^N, \{f^i\}_{i=1}^N)$, a set of strategies $\gamma^*$
are open-loop Nash equilibrium strategies if for every agent $i \in [N]$ and every
strategy $\gamma^i$, we have 
\vspace{-0.1cm}
\begin{align}\label{eq:Nash-definition}
    & J^i(x_0,\gamma^{*}) \leq   J^i(x_0,{\gamma^{i}},{\gamma^{-i*}}).
\end{align}
\end{definition}
Definition~\ref{def:Nash} implies that at equilibrium, no agent has any incentive for
changing its strategy and actions once it fixes the strategies and actions of all the
other agents to be their equilibrium strategies $\gamma^{-i*}$. However, finding Nash
equilibria is challenging, as solving~\eqref{eq:Nash-definition} requires solving a set
of coupled optimal control problems. Finding equilibria has proven difficult even in
two-player settings for discrete state and action spaces~\cite{fabrikant2004complexity,
ummels2008complexity, daskalakis2009complexity}. Most recent methods for finding Nash
equilibria of dynamic games that arise in robotics~\cite{fridovich2019ilqgames,
kavuncu2021potential, laine2021computation} solve the game in a centralized fashion,
i.e. each agent must maintain a full copy of all the other agents. As a result, the
existing methods and solvers are not scalable and become intractable beyond three agents
with simple dynamics. In this work, we seek to remedy this and develop a distributed
trajectory optimization algorithm for finding Nash equilibria of dynamic games
underlying multi-agent interactions.

%%%%%%%%%%%%%%%%%%%%%%%%%%%%%%%%%%%%%%%%%%%%%%%%%%%%%%%%%%%%%%%%%%%%%%%%%%%%%%%%%%%%%%%
\section{Prior Results} \label{sec:prior-work}

In this section, we review some of the results from our previous work that we will
utilize in the current work. Specifically, we review our previous results from
\cite{kavuncu2021potential} that allow one to bypass solving~\eqref{eq:Nash-definition}
for interactions that are dynamic potential games~\cite{fonseca2018potential}. More
specifically, we have the following result from~\cite{kavuncu2021potential}.

\begin{theorem} \label{thm:potential-dynamic-game}     
For a given dynamic game $G_{x_0}^T = (T, \{\gamma^i\}_{i=1}^{N}, \{J^i\}_{i=1}^{N},
\{f^i\}_{i=1}^{N})$, if for each agent $i \in [N]$, the running and terminal costs have
the the following structure
\begin{equation} \label{eqn:running-cost-thm}
    L^i(x_k, u_k) = p(x_k, u_k) + c^i(x_k^{-i}, u_k^{-i}),\; \forall k
\end{equation}
and
\begin{equation} \label{eqn:terminal-cost}
    S^i(x_T) = \bar{s}(x_T) + s^i(x^{-i}_T),
\end{equation}
then, the dynamic game $G_{x_0}^T$ is a dynamic potential game. and open-loop Nash
equilibria $u^* = (u^{1*}, \dots, u^{N*})$ can be found by solving the following optimal
control problem
\begin{equation} \label{eqn:single-optimal-control-problem}
\begin{split}
    & \min_u \sum_{k=0}^{T-1} p(x_k, u_k) + \bar{s}(x_T), \\
    & \text{s.t. } x_{k+1}^i = f^i(x_k^i, u_k^i).\\ % Q: Decoupled dynamics?
\end{split}
\end{equation}
\end{theorem}
Conditions~\eqref{eqn:running-cost-thm} and~\eqref{eqn:terminal-cost} imply that one can
decompose both the running costs $L^i(\cdot)$ and terminal costs $S^i(\cdot)$ into
potential functions $p(\cdot)$ and $\bar{s}(\cdot)$ which can depend on the full state
and control vector of the agents, and the cost terms $c^i(\cdot)$ and $s^i(\cdot)$ that
have no dependence on the state and control input of agent $i$.

For the remainder of the section, we make this result more concrete via a navigation
example, which serves as our running example throughout the paper. 

Consider a multi-agent navigation setup where each agent $i \in [N]$ must reach a goal
state $\bar{x}^i$. We assume that each agent's running cost function is composed of a
tracking cost and a control penalty term defined as:
\begin{align} \label{eqn:eg-tracking-cost}
    \begin{split} C^i_{tr} (x_k^i, u_k^i) &= (x_k^i - \bar{x}^i)\T Q^i (x_k^i - \bar{x}^i) \\
               &+ (u_k^i - \bar{u}^i) \T R^i (u_k^i - \bar{u}^i), \end{split} \\
    C^i_{tr, T} (x_T) & = (x^i_T - \bar{x}^i)\T Q_f^i (x^i_T - \bar{x}^i), \label{eqn:eg-term-cost}
\end{align}
where $Q^i, Q_f^i \in \R^{n_i \times n_i}$, $Q^i, Q_f^i \succeq 0$ and $R^i \in \R^{m_i
\times m_i}$, $R^i \succ 0$ are all symmetric matrices, and $\bar{u}^i$ is a reference
control input. Moreover, assume that agents' decisions are coupled through cost terms
such as the collision avoidance terms between any pair of agents $i\neq j, i, j \in [N]$
defined as:
\begin{equation} \label{eqn:eg-coupling-cost}
    C^{ij}_{ca} (x_k^i, x_k^j) = \alpha^{ij} (d(x_k^i, x_k^j)),
\end{equation}
where $d(x_k^i, x_k^j) \equiv d_k^{ij}$ is the distance between agents $i$ and $j$ at
time $k$. 

For each agent $i \in [N]$, the instantaneous running cost $L^i(\cdot)$ can then be
defined as:
\begin{equation}\label{eqn:running-cost}
    L^i(x_k, u_k^i) = C^i_{tr} (x_k^i, u_k^i) + \sum_{i \neq j}^{N} C^{ij}_{ca} (x_k^i, x_k^j).
\end{equation}

It was shown in \cite{kavuncu2021potential} that under the assumption that agents induce
coupling costs between each other symmetrically, i.e. $C^{ij}_{ca}(x_k^i, x_k^j) =
C^{ji}_{ca}(x_k^j, x_k^i), \forall i\neq j \in [N]$, the dynamic game is a dynamic
potential game with the potential function:
\begin{equation} \label{eqn:centralized-potential}
    p(x_k, u_k) = \sum_{i=1}^{N} C^i_{tr}(x^i_k, u^i_k) + \sum_{i=1, i<j}^{N} C^{ij}_{ca} (x^i_k, x^j_k).
\end{equation}
This implies that Nash Equilibria~\eqref{eq:Nash-definition} can be found by
solving~\eqref{eqn:single-optimal-control-problem}, which is a single optimal control
problem. Prior work~\cite{kavuncu2021potential} used iLQR~\cite{li2004ilqr,
tassa2012ilqr} for minimizing~\eqref{eqn:centralized-potential} in a Model Predictive
Control (MPC) setting. Explicitly, it solves~\eqref{eqn:single-optimal-control-problem}
for all agents iteratively as covered in Algorithm~\ref{alg:p-ilqr}. We refer the reader
to the original work~\cite{kavuncu2021potential} for additional details.
\begin{algorithm}
\caption{Potential-iLQR} \label{alg:p-ilqr}
\begin{algorithmic}
    \STATE \textbf{Inputs}
    \STATE dynamics $\{f^i\}_{i=1}^N$, potentials $p$ and $\bar{s}$, initial state $x_0$
    \STATE \textbf{Outputs}
    \STATE control inputs $u$
    \STATE \textbf{Initialization}
    \STATE $u \leftarrow 0^{N \times m}$
\end{algorithmic}
\algsetup{linenodelimiter=.}
\begin{algorithmic}[1]
    \WHILE{not converged}
        \STATE Rollout from $x_0$ with $u$ s.t.~\eqref{eqn:decoupled-dynamics} to compute $X$
        \STATE Linearize dynamics, quadraticize costs about $X, u$
        \STATE Solve the Riccati recursion to update controls $u$
    \ENDWHILE
\end{algorithmic}
\end{algorithm}
\vspace{-2ex}

%%%%%%%%%%%%%%%%%%%%%%%%%%%%%%%%%%%%%%%%%%%%%%%%%%%%%%%%%%%%%%%%%%%%%%%%%%%%%%%%%%%%%%%
\section{Distributed Interactive Trajectory Optimization} \label{sec:dist-traj-opt}
Prior work has proposed to solve~\eqref{eqn:centralized-potential} by requiring every
agent to maintain a copy of all the other agents, which restricts the scalability of the
method. Our key insight is that we can solve the optimal control
problem~\eqref{eqn:centralized-potential} more efficiently using ideas from distributed
MPC~\cite{camponogara2002distributed, zheng2017distributed,
keviczky2006decentralized,richards2004decentralized}. Specifically, we can break
up~\eqref{eqn:centralized-potential} into smaller subproblems more relevant to each
agent.

\begin{figure}
    \centering
    \begin{tikzpicture}
    \tikzstyle{every node}=[draw=black, shape=circle, fill= blue!20];
        \node (1) at (-1,-1){$1$};
        \node (2) at (0,0){$2$};
        \node (3) at (-2.5,0.){$3$};
        \node (4) at (1.5,1.4){$4$};
        \node (5) at (3.0, 0.1){$5$};
        \draw[ultra thick, blue!70] (1) -- (2);
        \draw[ultra thick, blue!70] (1) -- (3);
        \draw[ultra thick, blue!70] (2) -- (3);
        \draw[ultra thick, blue!70] (2) -- (4);
        \draw[ultra thick, blue!70] (4) -- (5);
    \end{tikzpicture}
    \caption{Schematic Example of an interaction graph for one-time step, where 
    $\mathcal{N}^1 = \{2, 3\}$, $\mathcal{N}^2 = \{1, 3, 4\}$, 
    $\mathcal{N}^3 = \{1, 2\}$, $\mathcal{N}^4 = \{2, 5\}$, and 
    $\mathcal{N}^5 = \{4\}$.}
    \label{fig:interaction-graph}
    \vspace{-2ex}
\end{figure}
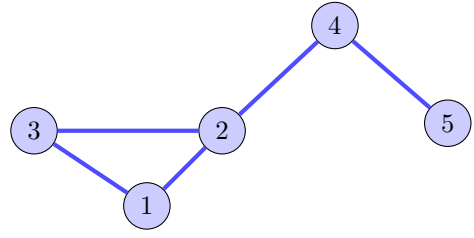

To formalize distributed interactive trajectory planning, we borrow ideas from
\cite{keviczky2006decentralized} and \cite{ferrari2009consensus} and introduce the
concept of an interaction graph $\mathcal{G} = \{ \mathcal{V}, \mathcal{E} \}$
comprising nodes $\mathcal{V} = [N]$ for each agent $i$ and edges $\mathcal{E} \in
\mathcal{V} \times \mathcal{V}$, where each edge $(i, j) \in \mathcal{V}$, indicates a
coupling between two agents in their costs. For our purposes, these edges are connected
if agents ever get within a proximity distance $d_{\text{prox}}$ of each other
throughout the predicted trajectories. Formally, let $\{ d^{ij}_k \}_{k=0}^{T-1} \in
\R^T$ be the predicted distances between agents $i$ and $j$ over the horizon $T$. We
create a bidirectional edge $(i, j)$ if:
\begin{equation} \label{eqn:interaction-graph-crit}
    \exists k , 0 \leq k < T, \; \text{such that} \; d^{ij}_k < \alpha d_{\text{prox}},
\end{equation}
where $\alpha \in \R, \alpha \geq 1$ is some aggressiveness parameter on how finely to
split up the graph. For sufficiently large $\alpha$, the interaction graph is complete,
i.e., all agents are connected with one another. Whereas for sufficiently small
$\alpha$, $\mathcal{E} = \emptyset$. Additionally, for each agent $i \in [N]$, its set
of neighbors is defined as $\mathcal{N}^i$, such that $\mathcal{N}^i = \{j: (i,j) \in
\mathcal{E}\}$ (see Fig.~\ref{fig:interaction-graph}). The introduction of the
interaction graph is motivated by human swarm motion. As referenced in
\cite{warren2018collective}, the key to collective motion is a pedestrian's zone of
influence. Intuitively, humans need not worry about others outside their vicinity.
Similarly, we argue that in multi-agent interactions, agents need only pay attention to
their zone of influence in the navigation problem. 

\begin{figure*}[ht]
    \centering
    \includegraphics[scale=1]{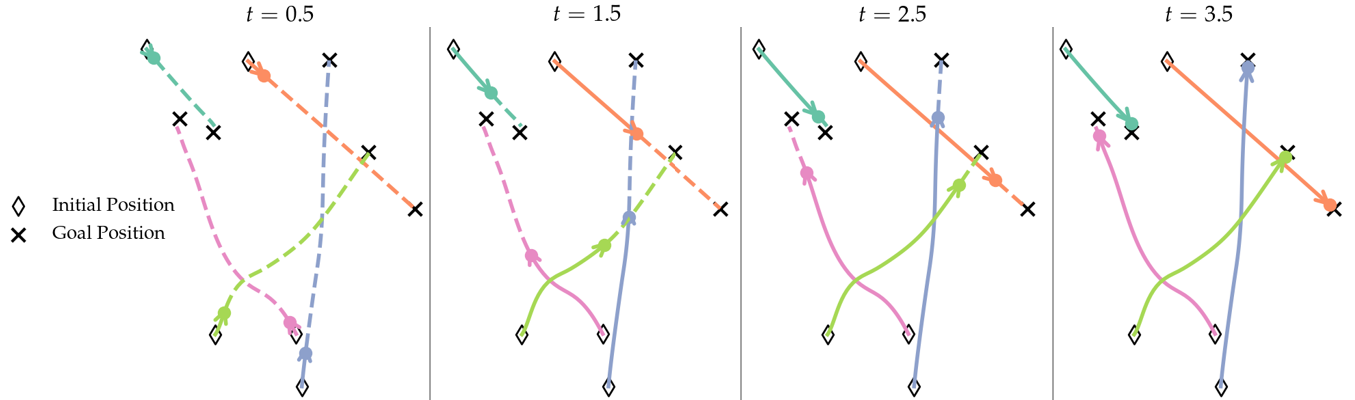}
    \caption{Example trajectories of DP-iLQR (Algorithm \ref{alg:dp-ilqr}) for 
    unicycle dynamics for varying numbers of agents from random initial conditions and 
    goal positions. As time progresses from the left to right, the algorithm is able 
    to guide agents through complex interactions.}
    \label{fig:example-trajectories}
    \vspace{-2ex}
\end{figure*}

We require each agent $i$ to solve its own subproblem, which we denote  by
$\mathcal{P}^i$, associated with minimizing the potential function that comprises only
itself and its neighbors $\tilde{p}^i$, i.e., the subset of the graph that it shares
edges with $\mathcal{N}^i$. Let the full subproblem consist of agents in the set
$\widetilde{\mathcal{N}}^i = \mathcal{N}^i \cup i$. The state vector of this subproblem
at time $k$ is then denoted by $\tilde{x}_k^i = \{x_k^j\}_{j \in
\widetilde{\mathcal{N}}^i}$ with the corresponding control input $\tilde{u}_k^i =
\{u_k^j\}_{j \in \widetilde{\mathcal{N}}^i}$. We propose to divide the centralized
problem~\eqref{eqn:single-optimal-control-problem} into a set of local subproblems,
where each agent $i$ solves its subproblem $\mathcal{P}^i$ defined as
\begin{equation} \label{eqn:dist-potential-problem}
\begin{split}
    & \min_{\tilde{u}^i} \sum_{k=0}^{T-1} \tilde{p}^i(\tilde{x}^i_k, \tilde{u}^i_k) + \tilde{\bar{s}}^i(\tilde{x}^i_T) \\
    & \text{s.t. } x^i_{k+1} = f^i (x^i_k, u^i_k), \\
\end{split}
\end{equation}
where $\tilde{\bar{s}}^i(\cdot) : \R^{\tilde{n}_i} \mapsto \R$ for $\tilde{n}_i =
\sum_{i \in \tilde{N}_i} n_i$, is a terminal cost for the local problem.
Therefore,~\eqref{eqn:dist-potential-problem} is the local analog
to~\eqref{eqn:single-optimal-control-problem} with local potential functions
$\tilde{p}^i$ and $\tilde{\bar{s}}^i$ that comprise local costs. In the specific case of
multi-agent navigation, one such potential function could take the following form:
\begin{equation} \label{eqn:distributed-potential}
    \tilde{p}^i(x_k, u_k) = \sum_{j \in \widetilde{\mathcal{N}}^i} C^j_{tr}(x^j_k, u^j_k) + \sum_{j \in \mathcal{N}^i} C^{ij}_{ca} (x^i_k, x^j_k).
\end{equation}
Hence,~\eqref{eqn:distributed-potential} is then a subset
of~\eqref{eqn:centralized-potential} that takes advantage of the sparsity of the
interaction graph.

Let the trajectory of the full system predicted by agent $i$ at time $k$ be $X^i_k$,
where $X^i_0 \equiv x_0$, such that over the full horizon:
\begin{equation} \label{eqn:predicted-trajectories}
    X^i = \{x_0, X^i_1, \dots, X^i_{T-1}\},
\end{equation}
where $X^i \in \R^{n \times T}$ is the predicted trajectory according to agent $i$
across the horizon. We are now ready to define our distributed trajectory planner in
Algorithm \ref{alg:dp-ilqr} --- Distributed Potential-iLQR (DP-iLQR).
\begin{algorithm}
    \caption{DP-iLQR}\label{alg:dp-ilqr}
    \begin{algorithmic}
        \STATE \textbf{Inputs}
        \STATE predicted trajectories $X^i$~\eqref{eqn:predicted-trajectories},
        system dynamics $\{f^j\}_{j=1}^N$~\eqref{eqn:decoupled-dynamics}, 
        costs~\crefrange{eqn:eg-tracking-cost}{eqn:eg-coupling-cost}
        \STATE \textbf{Outputs}
        \STATE control inputs $u^i$ for agent $i$
    \end{algorithmic}
    \algsetup{linenodelimiter=.}
    \begin{algorithmic}[1]
        \STATE $\mathcal{N}^i, \tilde{x}_0^i \leftarrow \text{define Interaction Graph}(X^i)$ ~\eqref{eqn:interaction-graph-crit} \label{alg:def-intergraph}
        \STATE $\tilde{p}^i, \tilde{\bar{s}}^i \leftarrow \text{define local potential functions}(\mathcal{N}^i)$ \label{alg:def-local}
        \STATE $U \leftarrow \text{Potential-iLQR}(\{f^i\}_{i \in \widetilde{\mathcal{N}}_i}, \tilde{p}^i, \tilde{\bar{s}}^i, \tilde{x}_0^i)$ (Alg.~\ref{alg:p-ilqr}) \label{alg:solve}
        \STATE $u^i \leftarrow$ \text{extract Agent}$(U, i)$ \label{alg:extract}
    \end{algorithmic}
\end{algorithm}
\vspace{-2ex}

In applying Algorithm \ref{alg:dp-ilqr} in a receding horizon fashion, we continually
compute the interaction graph at each step in line \ref{alg:def-intergraph} and compose
local potential functions in line \ref{alg:def-local}. Agents then solve their local
subproblem~\eqref{eqn:dist-potential-problem} in line \ref{alg:solve}. Agent $i$'s
solution $u^i$ is then pulled out from $U$ in line \ref{alg:extract} and executed. This
is then communicated with the rest of the system to define the subsequent graph at the
next step.

We can utilize any single-agent trajectory optimization algorithm to solve each local
subproblem by each agent $i$. We choose iLQR for solving each subproblem $\mathcal{P}^i$
due to its widespread success across many robotics domains. The broader question of how
these sub-problems interact with each other remains an open question, which we will
explore empirically in Section \ref{sec:simulation-studies}.

%%%%%%%%%%%%%%%%%%%%%%%%%%%%%%%%%%%%%%%%%%%%%%%%%%%%%%%%%%%%%%%%%%%%%%%%%%%%%%%%%%%%%%%
\section{SIMULATION STUDIES} \label{sec:simulation-studies} To evaluate the performance
of DP-iLQR, we execute the algorithm in a series of Monte Carlo simulations. We would
like to show that DP-iLQR can handle larger-sized problems than Potential-iLQR. To
accomplish this, we vary the number of agents and compare the centralized planner with
the distributed planner at a given initial condition. We considered a multi-agent
navigation setup with several different dynamics models, including 2D double integrator,
2D unicycle, and 3D quadcopter dynamics. We model our quadcopter as a six-dimensional
model, specifically:
\begin{equation} \label{eqn:quad-dynamics}
  \begin{aligned}
      \dot{p}_x & = v_x, & \dot{v}_x & = g \tan(\theta), \\
      \dot{p}_y & = v_y, & \dot{v}_y & = -g \tan(\phi),  \\
      \dot{p}_z & = v_z, & \dot{v}_z & = \tau - g,       \\
  \end{aligned}
\end{equation}
where $\theta$ is the pitch, $\phi$ is the roll, $\tau$ is the combined force of the
motors in the z-direction, $p_x$, $p_y$, and $p_z$ are the 3D position, $v_x$, $v_y$,
and $v_z$ are the 3D velocities, and $g$ is the acceleration due to gravity.

The specific form of collision avoidance cost that we utilized is the following:
\begin{equation} \label{eqn:eg-proximity-cost}
    \alpha^{ij} (d_k^{ij}) = \begin{cases}
        \beta (d_k^{ij} - d_{\text{prox}})^2 & d_k^{ij} < d_{\text{prox}} \\
        0                                  & \text{ otherwise},
    \end{cases}
\end{equation}
where $\beta \in \R$ is some weighting parameter and $d_{\text{prox}}$ is a threshold
distance where agents begin incurring penalties for being too close to each other.

We implemented Algorithm \ref{alg:dp-ilqr} in a combination of Python and C++. We
generated 30 random initial conditions. For a given initial condition, we fixed the
number of agents and computed the interactive trajectories of the agents using both
Potential-iLQR and DP-iLQR. We implemented our algorithm in a receding horizon
controller to fully exercise Algorithm \ref{alg:dp-ilqr}. We ran the simulations with a
horizon length of 40, a time interval of 0.1 seconds, and a collision radius
$d_{\text{prox}}$ of 0.5 meters. As shown in Fig.~\ref{fig:example-trajectories},
DP-iLQR generates intuitive collision-free trajectories under feasible initial
conditions.

We conducted two primary analyses to quantify the differences in the performance between
the centralized and distributed implementations: limited and unlimited solve time. The
solve time refers to the time it takes at a given operating point to entirely execute
the control algorithm, which corresponds to the time the algorithm would need to provide
a new control input in real-time use cases.

\subsection{Analysis 1 - Unlimited Solve Time}

\begin{figure}
    \centering
    \includegraphics[scale=0.5]{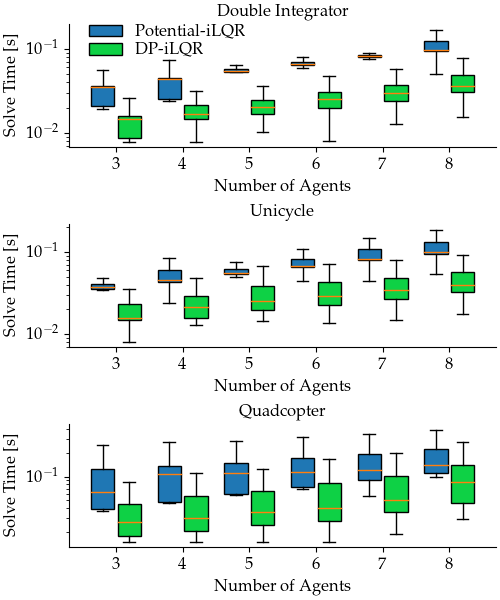}
    \caption{Average solve times without real-time constraints. DP-iLQR yields a 
        consistently lower solve time than the centralized solver across various 
        dynamics models and numbers of agents.}
    \label{fig:solve-times}
    \vspace{-2ex}
\end{figure}

For the case of unlimited solve time, we enforce no computational time limit on the
solver for converging to a solution at each receding horizon. This is an unrealistic
assumption for practical implementations, but it enables us to compare the solve time of
the two methods. In Fig. \ref{fig:solve-times}, we compare the individual subproblem
solve times throughout the simulated trajectories. As the number of agents increases,
the problem gets increasingly congested; hence, we see each agent spending more time
computing its subsequent control input. This is to be expected as the size of the state
space grows and the cost surface becomes more challenging to navigate. The advantage of
DP-iLQR in terms of the solve time is most prominent for simpler dynamical models like
the double integrator. Regardless, it demonstrates a consistent decrease in solve time
that widens with more agents.

\subsection{Analysis 2 - Real-Time Constraint}

\begin{figure}[ht]
    \centering
    \includegraphics[scale=0.5]{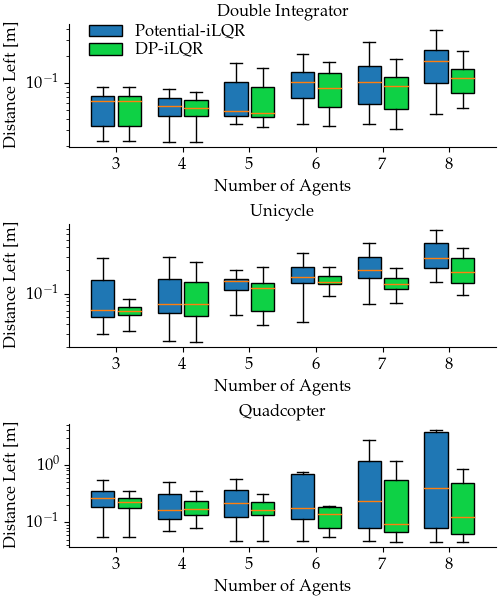}
    \caption{Average distance left to the goal position with a real-time constraint. 
        While the difference between the two solvers is smaller at lower scales, 
        Potential-iLQR is unable to maintain the quality of trajectories at the scales 
        of six or more agents.}
    \label{fig:dist-left}
    \vspace{-2ex}
\end{figure}

When evaluating performance with more realistic timing constraints, we only permit the
solver to iterate until it exceeds some time-based threshold. This serves as a
'best-effort' solution that will be more applicable to a hardware implementation where
the MPC planner requires a solution by the end of each time step. In this case, we
compare the quality of the trajectories under real-time constraints. In particular, we
measure the distance to goal at the end of the time horizon as a measure of solution
quality. Fig.~\ref{fig:dist-left} demonstrates the distance to the goal position under
the two methods for various numbers of agents.

As the number of agents exceeds six, we see in Fig. \ref{fig:dist-left} that the
variance in the centralized solver  increases significantly as it starts getting
overwhelmed handling the multitude of agent interactions. In contrast, the distributed
solver shows much more consistent convergence statistics and even outperforms the
centralized solver in most cases. 

%%%%%%%%%%%%%%%%%%%%%%%%%%%%%%%%%%%%%%%%%%%%%%%%%%%%%%%%%%%%%%%%%%%%%%%%%%%%%%%%%%%%%%%

\section{EXPERIMENTS} \label{sec:experiments}

To evaluate the real-time capabilities of our algorithm, we conducted navigation
experiments involving multiple quadcopters. We ran the DP-iLQR algorithm on five
Crazyflie 2.1 quadcopters, where each quadcopter navigates to its designated final
position. VICON motion capture system was used to provide position and velocity feedback
to all the quadcopters. The position update computed by DP-iLQR was sent to each
quadcopter online via the Crazyswarm API \cite{preiss2017crazyswarm}. The algorithm was
executed offboard on a laptop with an 8-core AMD processor with 32 GB of RAM. We
demonstrated that DP-iLQR provides a more stable and intuitive trajectory than the
Potential iLQR \cite{kavuncu2021potential}. A visualization demonstrating the
near-real-time trajectories generated by DP-iLQR is shown in Fig.
\ref{fig:5drone-experiment}.

%%%%%%%%%%%%%%%%%%%%%%%%%%%%%%%%%%%%%%%%%%%%%%%%%%%%%%%%%%%%%%%%%%%%%%%%%%%%%%%%%%%%%%%
\section{CONCLUSION}\label{sec:conclusion}

\textbf{Summary.}
We introduced a scalable and distributed algorithm for interactive trajectory planning
in multi-agent interactions. We considered interactions in a game-theoretic setting and
utilized the connection between multi-agent interactions and potential games to reduce
the problem of finding equilibria of the game to that of minimizing a single potential
function. Then, we used distributed trajectory optimization algorithms to minimize the
resulting potential function. This results in a scalable and efficient trajectory
planner for finding equilibria of interactive dynamic games. We compared our method with
the state-of-the-art in several simulation studies and hardware experiments involving
multiple quadcopters.

\textbf{Limitations and Future Work.}
In our simulations and experiments, we found that the consistency of the trajectory that
results from DP-iLQR is sensitive to the choice of cost parameters. More work would be
required to investigate the feasibility of designing a self-tuning algorithm to
determine optimal weights automatically. We would like to further examine both the
theoretical guarantees and optimality gap of our proposed trajectory planner, as well as
the application and impacts of distributed optimization techniques such as ADMM on
distributed interactive trajectory planning.

% NOTE: used to make the last page have even columns. Have to play with the length.
\addtolength{\textheight}{-3cm}

%%%%%%%%%%%%%%%%%%%%%%%%%%%%%%%%%%%%%%%%%%%%%%%%%%%%%%%%%%%%%%%%%%%%%%%%%%%%%%%%%%%%%%%
%%%%%%%%%%%%%%%%%%%%%%%%%%%%%%%%%%%%%%%%%%%%%%%%%%%%%%%%%%%%%%%%%%%%%%%%%%%%%%%%%%%%%%%

\bibliographystyle{ieeetr}
\bibliography{references}

\end{document}